\documentclass[letterpaper]{article} 
\usepackage{aaai24}  
\usepackage{times}  
\usepackage{helvet}  
\usepackage{courier}  
\usepackage[hyphens]{url}  
\usepackage{graphicx} 
\usepackage{natbib}  
\usepackage{caption} 
\usepackage{algorithm}
\usepackage{algorithmic}
\usepackage{times}
\usepackage{latexsym}

\usepackage[T1]{fontenc}

\usepackage[utf8]{inputenc}

\usepackage{microtype}

\usepackage{inconsolata}
\usepackage{algorithm}
\usepackage{algorithmic}
\usepackage{subfigure}
\usepackage{makecell}

\usepackage{multirow}
\usepackage{makecell}
\usepackage{subfigure}
\usepackage{multicol}
\usepackage{multirow}
\usepackage{url}
\usepackage{amsmath,amsfonts,amsthm} 
\usepackage{graphicx}
\usepackage{subfigure}
\usepackage{color}
\usepackage{tabularx}
\usepackage{array}
\usepackage{booktabs}
\usepackage{amssymb}
\usepackage{bbding}
\usepackage[subfigure]{tocloft}

\usepackage{newfloat}
\usepackage{listings}
\DeclareCaptionStyle{ruled}
{labelfont=normalfont,labelsep=colon,strut=off} 
\floatstyle{ruled}
\newfloat{listing}{tb}{lst}{}
\floatname{listing}{Listing}
\title{Learning to Rank in Generative Retrieval}
\author {
    Yongqi Li\textsuperscript{\rm 1},
    Nan Yang\textsuperscript{\rm 2},
    Liang Wang\textsuperscript{\rm 2},
    Furu Wei\textsuperscript{\rm 2},
    Wenjie Li\textsuperscript{\rm 1},
}
\affiliations {
    \textsuperscript{\rm 1}The Hong Kong Polytechnic University\\
    \textsuperscript{\rm 2}Microsoft\\
liyongqi0@gmail.com, \{nanya,wangliang,fuwei\}@microsoft.com, cswjli@comp.polyu.edu.hk
}

\usepackage{bibentry}

\begin{document}

\maketitle

\begin{abstract}
Generative retrieval stands out as a promising new paradigm in text retrieval that aims to \textit{generate} identifier strings of relevant passages as the retrieval target.  This generative paradigm taps into powerful generative language models, distinct from traditional sparse or dense retrieval methods. However, only learning to generate is insufficient for generative retrieval. Generative retrieval learns to generate identifiers of relevant passages as an intermediate goal and then converts predicted identifiers into the final passage rank list.  The disconnect between the learning objective of autoregressive models and the desired passage ranking target leads to a learning gap. To bridge this gap, we propose a learning-to-rank framework for generative retrieval, dubbed LTRGR. LTRGR enables generative retrieval to learn to rank passages directly, optimizing the autoregressive model toward the final passage ranking target via a rank loss. This framework only requires an additional learning-to-rank training phase to enhance current generative retrieval systems and does not add any burden to the inference stage. We conducted experiments on three public benchmarks, and the results demonstrate that LTRGR achieves state-of-the-art performance among generative retrieval methods. The code and checkpoints are released at \url{https://github.com/liyongqi67/LTRGR}. 
\end{abstract}

\section{Introduction}
Text retrieval is a crucial task in information retrieval and has a significant impact on various language systems, including search ranking~\cite{nogueira2019passage} and open-domain question answering~\cite{chen2017reading}. At its core, text retrieval involves learning a ranking model that assigns scores to documents based on a given query, a process known as \textit{learning to rank}. This approach has been enduringly popular for decades and has evolved into point-wise, pair-wise, and list-wise methods. Currently, the dominant implementation is the dual-encoder approach~\cite{lee2019latent,karpukhin2020dense}, which encodes queries and passages into vectors in a semantic space and employs a list-wise loss to learn the similarities.

An emerging alternative to the dual-encoder approach in text retrieval is generative retrieval~\cite{tay2022transformer, bevilacqua2022autoregressive}. Generative retrieval employs autoregressive language models to generate identifier strings of passages as an intermediate target for retrieval. An identifier is a distinctive string to represent a passage, such as Wikipedia titles to Wikipedia passages. The predicted identifiers are then mapped to ranked passages as the retrieval results. In this manner, generative retrieval treats passage retrieval as a standard sequence-to-sequence task, maximizing  the likelihood of the passage identifiers given the input query, distinct from previous learning-to-rank approaches.

There are two main approaches to generative retrieval regarding the identifier types. One approach, exemplified by the DSI system and its variants~\cite{tay2022transformer}, assigns a unique numeric ID to each passage, allowing predicted numeric IDs to directly correspond to passages on a one-to-one basis. However, this approach requires memorizing the mappings from passages to their numeric IDs, making it ineffective for large corpus sets.  The other approach~\cite{bevilacqua2022autoregressive} takes text spans from the passages as identifiers. While the text span-based identifiers are effective in the large-scale corpus, they no longer uniquely correspond to the passages. In their work, a heuristic-based function is employed to rank all the passages associated with the predicted identifiers. Following this line, \citeauthor{li2023multiview} proposed using multiview identifiers, which have achieved comparable results on commonly used benchmarks with large-scale corpus. In this work, we follow the latter approach to generative retrieval.

Despite its rapid development and substantial  potential, generative retrieval remains constrained. It relies on a heuristic function to convert predicted identifiers into a passage rank list, which requires sensitive hyperparameters and exists outside the learning framework. More importantly, generative retrieval generates identifiers as an intermediate goal rather than directly ranking candidate passages. This disconnect between the learning objective of generative retrieval and the intended passage ranking target brings a learning gap.  Consequently, even though the autoregressive model becomes proficient in generating accurate identifiers, the predicted identifiers cannot ensure an optimal passage ranking order.

Tackling the aforementioned issues is challenging, as they are inherent to the novel generative paradigm in text retrieval. However, a silver lining emerges from the extensive evolution of the adeptness learning-to-rank paradigm, which has demonstrated adeptness in optimizing the passage ranking objective. Inspired by this progress, we propose to enhance generative retrieval by integrating it with the classical learning-to-rank paradigm.
Our objective is to enhance generative retrieval to not solely generate fragments of passages but to directly acquire the skill of ranking passages. This shift aims to bridge the existing gap between the learning focus of generative retrieval and the envisaged passage ranking target.

\begin{figure*}[t]
\centering
  \includegraphics[width=1.0\linewidth]{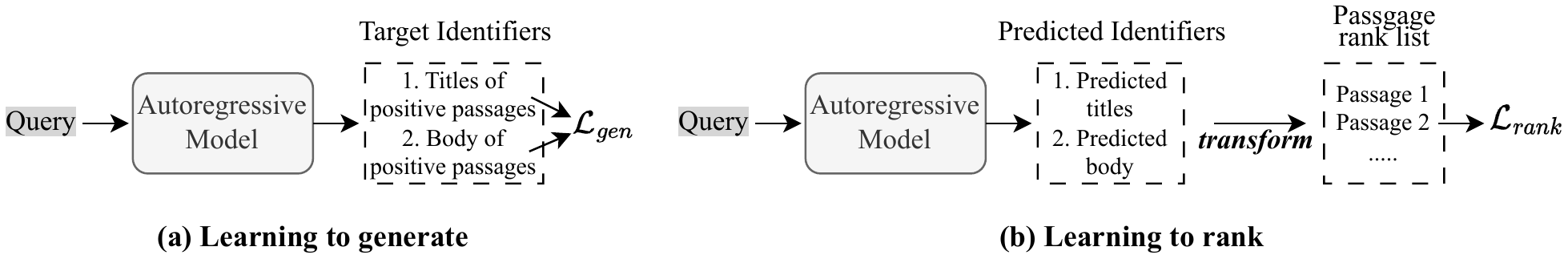}
  \vspace{-2em}
  \caption{This illustration depicts our proposed learning-to-rank framework for generative retrieval, which involves two stages of training. (a) Learning to generate: LTRGR first trains an autoregressive model via the generation loss, as a normal generative retrieval system. (b) Learning to rank: LTRGR continues training the model via the passage rank loss, which aligns the generative retrieval training with the desired passage ranking target.   }
    \vspace{-1em}
  \label{method}
\end{figure*}

In pursuit of this goal, we propose a learning-to-rank framework for generative retrieval, dubbed LTRGR. LTRGR involves two distinct training phases, as visually depicted in Figure~\ref{method}: the learning-to-generate phase and the learning-to-rank phase. In the initial learning-to-generate phase, we train an autoregressive model consistent with prior generative retrieval methods via the generation loss, which takes queries as input and outputs the identifiers of target passages. Subsequently, the queries from the training dataset are fed into the trained generative model to predict associated identifiers. These predicted identifiers are mapped to a passage rank list via a heuristic function. The subsequent learning-to-rank phase further trains the autoregressive model using a rank loss over the passage rank list, which optimizes the model towards the objective of the optimal passage ranking order. LTRGR includes the heuristic process in the learning process, rendering the whole retrieval process end-to-end and learning with the objective of passage ranking. During inference, we use the trained model to retrieve passages as in the typical generative retrieval. Therefore, the LTRGR framework only requires an additional training phase and does not add any burden to the inference stage. We evaluate our proposed method on three widely used datasets, and the results demonstrate that LTRGR achieves the best performance in generative retrieval.

The key contributions are summarized:
\begin{itemize}
\itemsep-0em 
  \item We introduce the concept of incorporating learning to rank within generative retrieval, effectively aligning the learning objective of generative retrieval with the desired  passage ranking target.
  \item LTRGR establishes a connection between the generative retrieval paradigm and the classical learning-to-rank paradigm. This connection opens doors for potential advancements in this area, including exploring diverse rank loss functions and negative sample mining.
    \item Only with an additional learning-to-rank training phase and without any burden to the inference, LTRGR achieves state-of-the-art performance in generative retrieval on three widely-used benchmarks. 
\end{itemize}

\section{Related Work}
\subsection{Generative Retrieval}
Generative retrieval is an emerging new retrieval paradigm, which generates identifier strings of passages as the retrieval target. Instead of generating entire passages, this approach uses identifiers to reduce the amount of useless information and make it easier for the model to memorize and learn~\cite{li2023multiview}. Different types of identifiers have been explored in various search scenarios, including titles (Web URLs), numeric IDs, and substrings, as shown in previous studies~\cite{de2020autoregressive, li2023generative, tay2022transformer,bevilacqua2022autoregressive,ren2023tome}. In 2023, ~\citeauthor{li2023multiview} proposed multiview identifiers that represented a passage from different perspectives to enhance generative retrieval and achieve state-of-the-art performance. Despite the potential advantages of generative retrieval, there are still issues inherent in this new paradigm, as discussed in the previous section. Our work aims to address these issues by combining generative retrieval with the learning-to-rank paradigm.

\subsection{Learning to Rank}
Learning to rank refers to machine learning techniques used for training models in ranking tasks~\cite{li2011short}. This approach has been developed over several decades and is typically applied in document retrieval. Learning to rank can derive large-scale training data from search log data and automatically create the ranking model, making it one of the key technologies for modern web search. Learning to rank approaches can be categorized into point-wise~\cite{cossock2006subset,li2007mcrank,crammer2001pranking}, pair-wise~\cite{freund2003efficient,burges2005learning}, and list-wise~\cite{cao2007learning,xia2008listwise} approaches based on the learning target. In the point-wise and pair-wise approaches, the ranking problem is transformed into classification and pair-wise classification, respectively. Therefore, the group structure of ranking is ignored in these approaches. The list-wise approach addresses the ranking problem more directly by taking ranking lists as instances in both learning and prediction. This approach maintains the group structure of ranking, and ranking evaluation measures can be more directly incorporated into the loss functions in learning.
\subsection{Dense Retrieval}
Dense retrieval~\cite{lee2019latent,karpukhin2020dense}, which is an extension of learning to rank in the context of large language models, is currently the de facto implementation of document retrieval. This method benefits from the powerful representation abilities of large language models and the MIPS algorithm~\cite{shrivastava2014asymmetric}, allowing for efficient passage retrieval from a large-scale corpus. Dense retrieval has been further developed through hard negative sample mining~\cite{xiong2020approximate,qu2021rocketqa,li2022dynamic} and better pre-training design~\cite{chang2019pre,wang2022simlm}, resulting in an excellent performance. However, compared to dense retrieval, which relies on the dual-encoder architecture, generative retrieval shows promise in overcoming the missing fine-grained interaction problem through the encoder-decoder paradigm. Despite being a recently proposed technique, generative retrieval still lags behind the state-of-the-art dense retrieval method and leaves much room for investigation. In this work, we introduce a promising way to further develop generative retrieval systems.
\section{Method}
When given a query text $q$, the retrieval system must retrieve a list of passages $\{p_1, p_2, \dots, p_n\}$ from a corpus $\mathcal{C}$, where both queries and passages consist of a sequence of text tokens. As illustrated in Figure~\ref{method}, LTRGR involves two training stages: learning to generate and learning to rank. In this section, we will first provide an overview of how a typical generative retrieval system works. i.e. learning to generate, and then clarify our learning-to-rank framework within the context of generative retrieval.
\subsection{Learning to Generate}
We first train an autoregressive language model using the standard sequence-to-sequence loss. In practice, we follow the current sota generative retrieval method, MINDER~\cite{li2023multiview}, to train an autoregressive language model. Please refer to the MINDER for more details.

\textbf{Training}. 
We develop an autoregressive language model, referred to as $\textbf{AM}$, to generate multiview identifiers. The model takes as input the query text and an identifier prefix, and produces a corresponding identifier of the relevant passage as output. The identifier prefix can be one of three types: "title", "substring", or "pseudo-query", representing the three different views. The target text for each view is the title, a random substring, or a pseudo-query of the target passage, respectively. During training, the three different samples are randomly shuffled to train the autoregressive model.

For each training sample,  the objective is to minimize the sum of the negative loglikelihoods of the tokens $\{i_1,\cdots,i_j,\cdots,i_l\}$ in a target identifier $I$, whose length is $l$. The generation loss is formulated as,
\begin{equation}  \label{eqn1}
   \begin{aligned}
   &\mathcal{L}_{gen} = 
   &-\sum_{j=1}^{l}\log p_{\theta}({i_j}|q;I_{<j}),
   \end{aligned}
 \end{equation}
where $I_{<j}$ denotes the partial identifier sequence $\{i_0,\cdots,i_{j-1}\}$, $i_0$ is a pre-defined start token, and $\theta$ is the trainable parameters in the autoregessive model $\textbf{AM}$.

\textbf{Inference}. During the inference process, given a query text, the trained autoregressive language model $\textbf{AM}$ could generate predicted identifiers in an autoregressive manner.  The FM-index~\cite{892127} data structure is used to support generating valid identifiers. Given a start token or a string, FM-index could provide the list of possible token successors. Therefore, we could store all identifiers of passages in $\mathcal{C}$ into FM-index and thus force the $\textbf{AM}$ model to generate valid identifiers via constrained generation. Given a query $q$, we could set different identifier prefixes to generate a series of predicted identifiers $\mathcal{I}$ via beam search, formulated as,
\begin{equation}  \label{eqn2}
   \begin{aligned}
   \mathcal{I} = \textbf{AM}(q; b; \textit{FM-index}),
   \end{aligned}
 \end{equation}
where $b$ is the beam size for beam search.

In order to retrieve passages from a large corpus, a heuristic function is employed to transform the predicted identifiers $\mathcal{I}$ into a ranked list of passages. We give a simple explanation, and please refer to the original paper for details. For each passage $p \in \mathbf{C}$, we select a subset $\mathcal{I}_p$ from the predicted
identifiers $\mathcal{I}$, where  $i_p \in \mathcal{I}_p$ if $i_p$ is one of the identifiers of the passage $p$. The rank score of the passage $p$ corresponding to the query $q$ is then calculated as the sum of the scores of its covered identifiers,
\begin{equation}  \label{eqn3}
   \begin{aligned}
   s(q, p) = \sum_{i_p \in \mathcal{I}_p} s_{i_p},
   \end{aligned}
 \end{equation}
where $s_{i_p}$ represents the language model score of the identifier $i_p$, and $\mathcal{I}_p$ is the set of selected identifiers that appear in the passage $p$. By sorting the rank score $s(q, p)$, we are able to obtain a ranked list of passages from the corpus $\mathcal{C}$. In practice, we can use the FM-index to efficiently locate those passages that contain at least one predicted identifier, rather than scoring all of the passages in the corpus.

\subsection{Learning to Rank}
As previously mentioned, it is insufficient for generative retrieval to only learn how to generate identifiers. Therefore, we develop a framework to enable generative retrieval to learn how to rank passages directly. To accomplish this, we continue training the autoregressive model $\textbf{AM}$ using a passage rank loss.

To begin, we retrieve passages for all queries in the training set using the trained autoregressive language model $\textbf{AM}$ after the learning-to-generate phase. For a given query $q$, we obtain a passage rank list $\mathcal{P}=\{p_1,\cdots,p_j,\cdots, p_n\}$, where $n$ is the number of retrieved passages. Each passage $p_j$ is assigned a relevant score $s(q, p_j)$ via Eq.~\ref{eqn3}, which is calculated as the sum of the language model scores of a set of predicted identifiers. It is important to note that the passage rank list includes both positive passages that are relevant to the query and negative passages that are not.

\begin{table*}[t]
\renewcommand\arraystretch{1}
  \centering
    \scalebox{1.0}{
    \begin{tabular}{cccccccc}
    \toprule
    \multicolumn{1}{c}{\multirow{2}*{Methods}}
    &\multicolumn{3}{c}{\makecell[c]{Natural Questions}}&&\multicolumn{3}{c}{\makecell[c]{TriviaQA}}\\\cline{2-4}\cline{6-8}
         &@5&@20&@100&&@5&@20&@100\cr
    \toprule
    BM25&43.6&62.9&78.1&&67.7&77.3&83.9\cr
    DPR\cite{karpukhin2020dense}&\underline{68.3}&\underline{80.1}&86.1&&\underline{72.7}&\underline{80.2}&{84.8}\cr 
    GAR\cite{mao-etal-2021-generation}&59.3&73.9&85.0&&\textbf{73.1}&\textbf{80.4}&\textbf{85.7}\cr \toprule
    DSI-BART\cite{tay2022transformer}&28.3&47.3&65.5&&-&-&-\cr 
    SEAL-LM\cite{bevilacqua2022autoregressive}&40.5&60.2&73.1&&39.6&57.5&80.1\cr 
    SEAL-LM+FM\cite{bevilacqua2022autoregressive}&43.9&65.8&81.1&&38.4&56.6&80.1\cr SEAL\cite{bevilacqua2022autoregressive}&61.3&76.2&{86.3}&&66.8&77.6&84.6\cr 
    MINDER\cite{li2023multiview}&${65.8}$&${78.3}$&$\underline{86.7}$&&$68.4$&$78.1$&${84.8}$\cr
LTRGR&$\textbf{68.8}^{\dagger}$&$\textbf{80.3}^{\dagger}$&$\textbf{87.1}^{\dagger}$&&$70.2^{\dagger}$&$79.1^{\dagger}$&$\underline{85.1}^{\dagger}$\cr
\% \textbf{improve}&4.56\%&2.55\%&0.46\%&&2.63\%&1.28\%&0.35\%\cr
    \toprule
    \end{tabular}} 
    \vspace{-0.5em}
    \caption{ Retrieval performance on NQ and TriviaQA. We use hits@5, @20, and @100, to evaluate the retrieval performance. Inapplicable results are marked by “-”. The best results in each group are marked in Bold, while the second-best ones are underlined. \textbf{$\dagger$ denotes the best result in generative retrieval}. \% improve represents the relative improvement achieved by LTRGR over the previously best generative retrieval method.}  \label{tab:Retrieval performance}
    \vspace{-0.5em}
\end{table*}
\begin{table*}[t]
\renewcommand\arraystretch{1}
  \centering
    \scalebox{1.0}{
    \begin{tabular}{cccccc}
    \toprule
    \multicolumn{1}{c}{\multirow{2}*{Methods}}&\multicolumn{1}{c}{\multirow{2}*{Model Size}}
    &\multicolumn{4}{c}{\makecell[c]{MSMARCO}}\\\cline{3-6}
         &&R@5&R@20&R@100&M@10\cr
    \toprule
    BM25&-&28.6&47.5&66.2&18.4\cr
SEAL\cite{bevilacqua2022autoregressive}&BART-Large&19.8&35.3&57.2&12.7\cr
    MINDER\cite{li2023multiview}&BART-Large&{29.5}&{53.5}&{78.7}&{18.6}\cr
    NCI\cite{wang2022neural}&T5-Base&{-}&{-}&{-}&{9.1}\cr
    DSI(scaling up)\cite{pradeep2023does}&T5-Base&{-}&{-}&{-}&{17.3}\cr
    DSI(scaling up)\cite{pradeep2023does}&T5-Large&{-}&{-}&{-}&{19.8}\cr
    LTRGR&BART-Large&\textbf{40.2}&\textbf{64.5}&\textbf{85.2}&\textbf{25.5}\cr
    \% \textbf{improve}&-&36.3\%&20.6\%&8.26\%&28.8\%\cr
    \toprule
    \end{tabular}}  
    \vspace{-0.5em}
    \caption{Retrieval performance on the MSMARCO dataset. R and M denote Recall and MRR, respectively.  “-” means the result  not reported in the published work. The best results in each group are marked in Bold. \% improve represents the relative improvement achieved by LTRGR over the previously best generative retrieval method. }  \label{tab:search dataset}
    \vspace{-1em}
\end{table*}

A reliable retrieval system should assign a higher score to positive passages than to negative passages, which is the goal of the learning-to-rank paradigm. To achieve this objective in generative retrieval, we utilize a margin-based rank loss, which is formulated as follows:
\begin{equation}  \label{eqn4}
   \begin{aligned}
   \mathcal{L}_{rank} =max(0,s(q,p_n)-s(q,p_p)+m),
   \end{aligned}
 \end{equation}
where $p_p$ and $p_n$ represent a positive and negative passage in the list $\mathcal{P}$, respectively, and $m$ is the margin. It is noted that the gradients could be propagated to the autoregressive model $\textbf{AM}$ via the language model score $s_{i_p}$, which is the logits of the neural network.

In practice, we take two rank losses based on different sampling strategies for positive and negative passages. In $\mathcal{L}_{rank1}$, the positive and negative passages are the ones with the highest rank scores, respectively. In $\mathcal{L}_{rank2}$, both the positive and negative passages are randomly sampled from the passage rank list. While the rank loss optimizes the autoregressive model toward passage ranking, the generation of identifiers is also crucial for successful passage ranking. Therefore, we also incorporate the generation loss into the learning-to-rank stage. The final loss is formulated as a multi-task format:
\begin{equation}  \label{eqn5}
   \begin{aligned}
   \mathcal{L} =\mathcal{L}_{rank1}+\mathcal{L}_{rank2}+\lambda\mathcal{L}_{gen},
   \end{aligned}
 \end{equation}
where $\lambda$ is the weight to balance the rank losses and generation loss.

We continue training the autoregressive model $\textbf{AM}$ via Eq.~\ref{eqn5}. After training, $\textbf{AM}$ can be used to retrieve passages as introduced in the learning to generate section. Therefore, our learning-to-rank framework does not add any additional burden to the original inference stage.

\section{Experiments}
\subsection{Datasets}
We conducted experiments using the DPR~\cite{karpukhin2020dense} setting on two widely-used open-domain QA datasets: NQ~\cite{kwiatkowski2019natural} and TriviaQA~\cite{joshi2017triviaqa}. In both datasets, the queries are natural language questions and the passages are sourced from Wikipedia. Additionally, we evaluated generative retrieval methods on the MSMARCO dataset~\cite{nguyen2016ms}, which is sourced from the Web search scenario where queries are web search queries and passages are from web pages. Importantly, we evaluated models on the full corpus set rather than a small sample, and we used widely-used metrics for these benchmarks.

\subsection{Baselines} 
We compared LTRGR with several generative retrieval methods, including DSI~\cite{tay2022transformer}, DSI (scaling up)~\cite{pradeep2023does}, NCI~\cite{wang2022neural}, SEAL~\cite{bevilacqua2022autoregressive}, and MINDER~\cite{li2023multiview}. Additionally, we included the term-based method BM25, as well as DPR~\cite{karpukhin2020dense} and GAR~\cite{mao-etal-2021-generation}. All baseline results were obtained from their respective papers.

\subsection{Implementation Details}
To ensure a fair comparison with previous work, we utilized BART-large as our backbone. In practice, we loaded the trained autoregressive model, MINDER~\cite{li2023multiview}, and continued training it using our proposed learning-to-rank framework. In the learning to rank phase, we used the Adam optimizer with a learning rate of 1e-5, trained with a batch size of 4, and conducted training for three epochs. For each query in the training set, we retrieved the top 200 passages and selected positive and negative passages from them. During training, we kept 40 predicted identifiers for each passage and removed any exceeding ones. The margin $m$ and weight $\lambda$ are set as 500 and 1000, respectively. Our main experiments were conducted on a single NVIDIA A100 GPU with 80 GB of memory.

\subsection{Retrieval Results on QA}
Table~\ref{tab:Retrieval performance} summarizes the retrieval performance on NQ and TriviaQA. By analyzing the results, we discovered the following findings:

(1) Among the generative retrieval methods, we found that SEAL and MINDER, which use semantic identifiers, outperform DSI, which relies on numeric identifiers. This is because numeric identifiers lack semantic information, and DSI requires the model to memorize the mapping from passages to their numeric IDs. As a result, DSI struggles with datasets like NQ and TriviaQA, which contain over 20 million passages. MINDER surpasses SEAL by using multiview identifiers to represent a passage more comprehensively. Despite MINDER's superiority, LTRGR still outperforms it. Specifically, LTRGR improves hits@5 by 3.0 and 1.8 on NQ and TriviaQA, respectively. LTRGR is based on MINDER and only requires an additional learning-to-rank phase, which verifies the effectiveness of learning to rank in generative retrieval.

(2) Regarding the NQ dataset, MINDER outperforms the classical DPR and achieves the best performance across all metrics, including hits@5, 20, and 100. This is particularly noteworthy as it marks the first time that generative retrieval has surpassed DPR in all metrics under the full corpus set setting.
Turning to TriviaQA, our results show that LTRGR outperforms DPR in hits@100, but falls behind in hits@5 and hits@20. The reason for this is that MINDER, upon which LTRGR is based, performs significantly worse than DPR on TriviaQA. It's worth noting that generative retrieval methods rely on identifiers and cannot "see" the content of the passage, which may explain the performance gap between MINDER and DPR on TriviaQA. Additionally, generative retrieval methods have an error accumulation problem in an autoregressive generative way.

\subsection{Retrieval Results on Web Search}
To further investigate generative retrieval, we conducted experiments on the MSMARCO dataset and presented our findings in Table~\ref{tab:search dataset}. It's worth noting that we labeled the model sizes to ensure a fair comparison, as larger model parameters typically result in better performance.

Our analysis of the results in Table~\ref{tab:search dataset} revealed several key findings. Firstly, we observed that generative retrieval methods perform worse in the search scenario compared to the QA datasets. Specifically, SEAL, NCI, and DSI underperformed BM25, while MINDER and DSI (T5-large) only slightly outperformed BM25. This is likely due to the fact that the passages in MSMARCO are sourced from the web, and are therefore of lower quality and typically lack important metadata such as titles.
Secondly, we found that LTRGR achieved the best performance and outperformed all baselines significantly. LTRGR surpassed the second-best approach, DSI (scaling up), by 5.7 points in terms of MRR@10, despite DSI using the larger T5-Large backbone compared to BART-Large.
Finally, we observed that the learning-to-rank paradigm significantly improves existing generative retrieval methods in the search scenario. Specifically, LTRGR improved MINDER by 10.7 points and 6.9 points in terms of Recall@5 and MRR@10, respectively. These results provide strong evidence of the effectiveness of LTRGR, which only requires an additional training step on MINDER.

\begin{table}[t]
\renewcommand\arraystretch{1}
  \centering
    \scalebox{1.0}{
    \begin{tabular}{cccc}
    \toprule
    \multicolumn{1}{c}{\multirow{2}*{Methods}}
    &\multicolumn{3}{c}{\makecell[c]{Natural Questions}}\\\cline{2-4}
         &@5&@20&@100\cr
    \toprule
    w/o learning-to-rank&65.8&78.3&86.7\cr\toprule
    w/ rank loss 1&56.1&69.4&78.7\cr
    w/o generation loss&63.9&76.1&84.4\cr
    w/o rank loss &65.8&78.6&86.5\cr
    w/o rank loss 1&68.2&80.8&87.0\cr
    w/o rank loss 2&67.9&79.8&86.7\cr\toprule
    LTRGR&68.8&80.3&87.1\cr \toprule
    \end{tabular}}  
    \vspace{-0.5em}
    \caption{ Ablation study of LTRGR with different losses in the learning-to-rank training phase. ``w/o learning-to-rank'' refers to the basic generative retrieval model, MINDER, without the learning-to-rank training.} 
    \label{tab:ablation study}
    \vspace{-1em}
\end{table}

\subsection{Ablation Study}
The LTRGR model is trained by leveraging the MINDER model and minimizing the loss function defined in Eq.~\ref{eqn5}. This loss function consists of two margin-based losses and one generation loss. To shed light on the role of the learning-to-rank objective and the impact of the margin-based losses, we conducted experiments where we removed one or more terms from the loss function. Specifically, we investigated the following scenarios:
\begin{itemize}
\itemsep-0em
\item ``w/o generation loss'': We removed the generation loss term ($\mathcal{L}_{gen}$) from the loss function, which means that we trained the autoregressive model solely based on the rank loss.
\item ``w/o rank loss'': We removed both margin-based losses ($\mathcal{L}_{rank1}$ and $\mathcal{L}_{rank2}$) from the loss function, which means that we trained the autoregressive model solely based on the generation loss, following a common generative retrieval approach.
\item ``w/o rank loss 1'' and ``w/o rank loss 2'': We removed one of the margin-based losses ($\mathcal{L}_{rank1}$ or $\mathcal{L}_{rank2}$) from the loss function, respectively.
\end{itemize}
Our experiments aimed to answer the following questions: Does the performance improvement of the LTRGR model come from the learning-to-rank objective or from continuous training? Is it necessary to have two margin-based losses? What happens if we train the model only with the rank loss?

We present the results of our ablation study in Table~\ref{tab:ablation study}, which provide the following insights:
(1) Removing the rank loss and training the model solely based on the generation loss does not significantly affect the performance. This observation is reasonable since it is equivalent to increasing the training steps of a generative retrieval approach. This result confirms that the learning-to-rank objective is the primary source of performance improvement and validates the effectiveness of our proposed method.
(2) Removing either $\mathcal{L}_{rank1}$ or $\mathcal{L}_{rank2}$ leads to a drop in the performance of LTRGR. On the one hand, having two rank losses allows the model to leverage a larger number of passages and benefits the rank learning. On the other hand, the two rank losses adopt different sample mining strategies, ensuring the diversity of the passages in the loss.
(3) Removing the generation loss is the only variant underperforming the original MINDER model.  During our experiments, we observed that the model tends to fall into local minima and assign smaller scores to all passages. This finding suggests the necessity of the generation loss in the learning-to-rank phase.
(4) Overall, the current loss function is the best choice for the learning-to-rank phase. We also explore the list-wise rank loss in Section 4.7.

\begin{table}[t]
\renewcommand\arraystretch{1}
  \centering
    \scalebox{1.0}{
    \begin{tabular}{cccc}
    \toprule
    \multicolumn{1}{c}{\multirow{2}*{Methods}}
    &\multicolumn{3}{c}{\makecell[c]{Natural Questions}}\\\cline{2-4}
         &@5&@20&@100\cr
    \toprule
    SEAL &61.3&76.2&{86.3}\cr
    SEAL-LTR&63.7&78.1&86.4\cr\toprule
    \end{tabular}}  
    \vspace{-0.5em}
    \caption{ Retrieval performance of SEAL and SEAL-LTR on NQ. SEAL-LTR represents applying our proposed LTRGR framework to the SEAL model.} 
    \label{tab:SEAL-LTR}
\end{table}

\begin{figure}[t!]
\centering
\subfigure[] {
\includegraphics[width=0.47\linewidth]{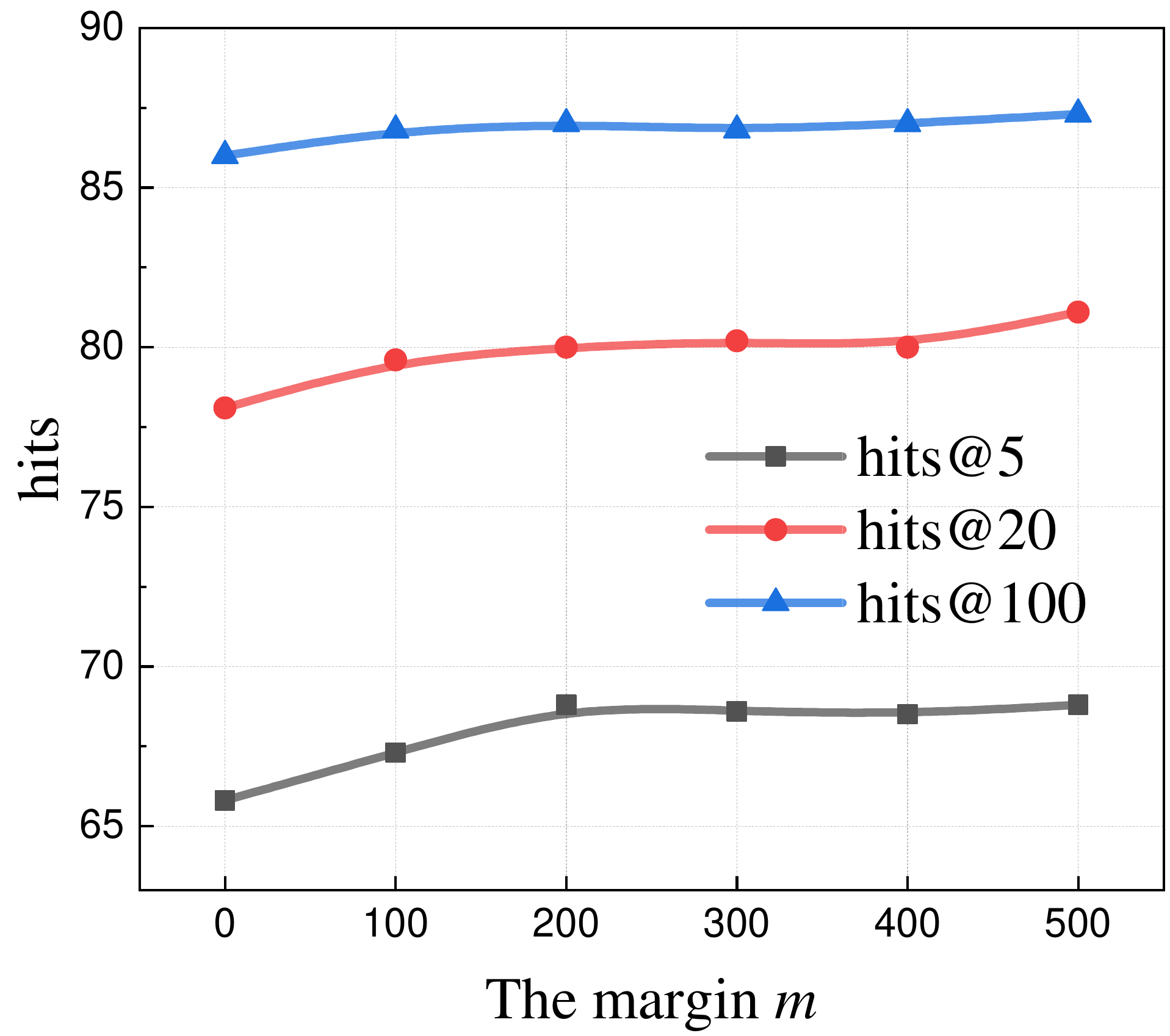}
   \label{margin_lambda_analysis:subfig1}
   }
\subfigure[] {
  \includegraphics[width=0.47\linewidth]{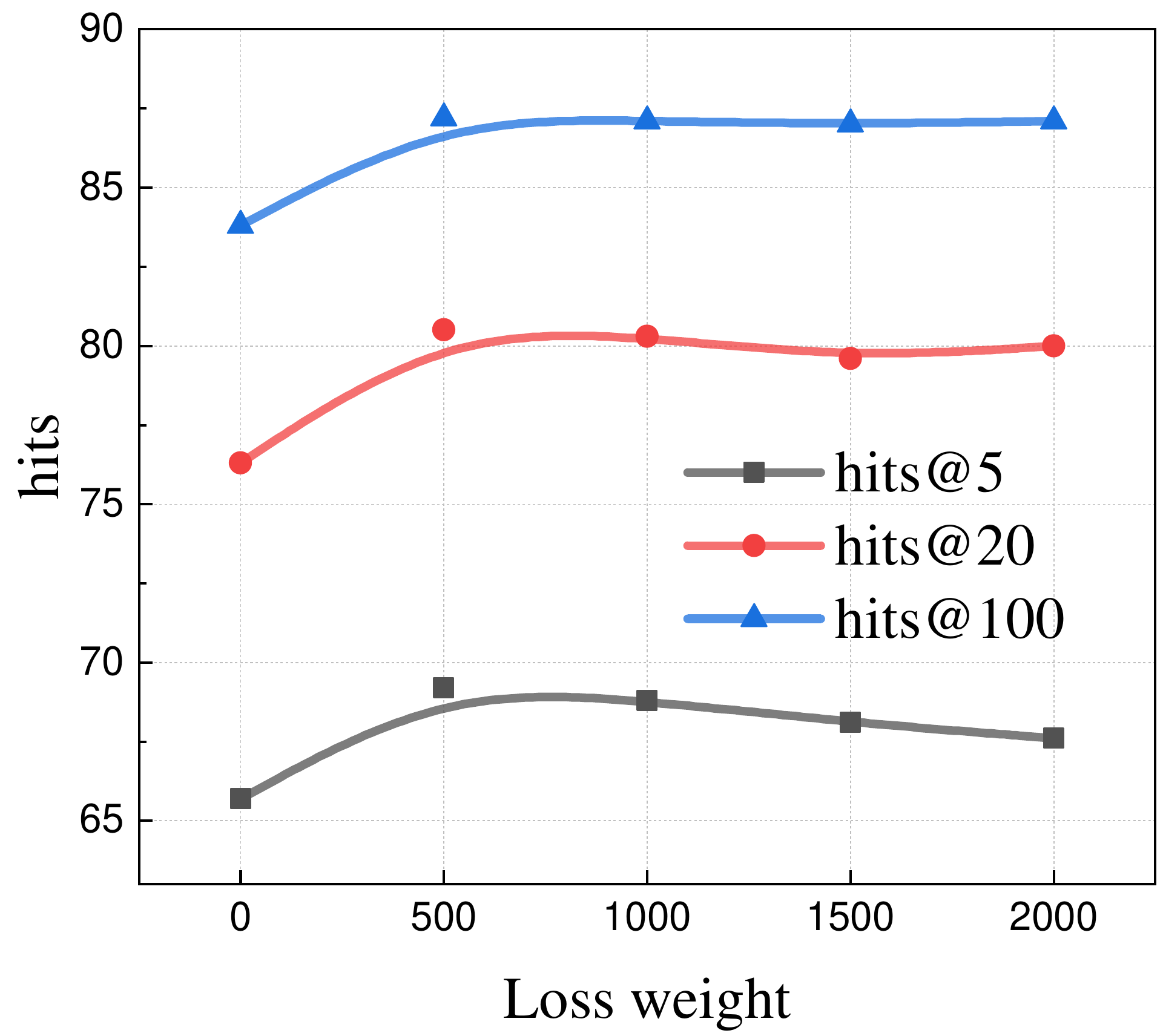}
   \label{margin_lambda_analysis:subfig2}
   }
\vspace{-1em}
\caption{ The retrieval performances of LTRGR on the NQ test set are shown in (a) and (b) against the margin values and balance weight $\lambda$, respectively.}
\vspace{-1em}
\label{margin_lambda_analysis}
\end{figure}

\subsection{In-depth Analysis}
\textbf{Generalization of LTRGR}. Our LTRGR builds on the generative retrieval model MINDER and continues to train it using the loss function described in Eq.~\ref{eqn5}. A natural question arises: can LTRGR be generalized to other generative retrieval models? To answer this question, we replaced MINDER with SEAL as the basic model and performed the same learning-to-rank training. The results, presented in Table~\ref{tab:SEAL-LTR}, show that the proposed LTRGR framework can also improve the performance of SEAL. Specifically, the hits@5, 20, and 100 metrics improved by 3.6, 1.9, and 0.1 points, respectively. Interestingly, we observed that the improvement on hits@5 was larger than that on hits@100, which may be attributed to the optimization of the top ranking using $\mathcal{L}_{rank1}$.

\textbf{List-wise loss}. To facilitate generative retrieval learning to rank, we adopt a margin-based loss as the rank loss. By doing so, LTRGR effectively connects generative retrieval with the learning-to-rank paradigm, allowing for various types of rank loss to be applied. To examine the impact of different rank losses, we substitute the original margin-based loss with a list-wise loss known as infoNCE, which is formulated as follows: 
\begin{equation}  \label{eqn6}
   \begin{aligned}
   &\mathcal{L}_{rank} = 
   &-\log \frac{\textbf{e}^{s(q,p_p)}} {\textbf{e}^{s(q,p_p)}+\sum_{p_n} \textbf{e}^{s(q,p_n)} }.
   \end{aligned}
 \end{equation}
We randomly selected 19 negative passages from the passage rank list $\mathcal{P}$ and presented the results in Table~\ref{tab:list-wise}.  It was observed that LTRGR with the infoNCE loss performed worse than the model with the margin-based loss. There are two potential reasons: Firstly, we only trained the model for one epoch due to the increased training cost, which may have resulted in insufficient training. Secondly, the passage scores were not normalized, making them difficult to optimize. 
The results also indicate that more suitable list-wise learning methods should be developed in generative retrieval.
\begin{table}[t]
\renewcommand\arraystretch{1}
  \centering
    \scalebox{1.0}{
    \begin{tabular}{cccc}
    \toprule
    \multicolumn{1}{c}{\multirow{2}*{Rank loss}}
    &\multicolumn{3}{c}{\makecell[c]{Natural Questions}}\\\cline{2-4}
         &@5&@20&@100\cr
    \toprule
    Margin loss &68.8&80.3&87.1\cr
    List-wise loss&65.4&78.5&86.3\cr\toprule
    \end{tabular}}  
    \vspace{-0.5em}
    \caption{ Performance comparison of LTRGR with the margin-based loss and the list-wise loss.} 
     \vspace{-1em}
    \label{tab:list-wise}
\end{table}

\textbf{Inference speed}. LTRGR simply adds an extra training step to existing generative models, without affecting inference speed. The speed of inference is determined by the underlying generative retrieval model and the beam size. We conducted tests on LTRGR using a beam size of 15 on one V100 GPU with 32GB memory. On the NQ test set, LTRGR based on MINDER took approximately 135 minutes to complete the inference process, while LTRGR based on SEAL took only 115 minutes. Notably, SEAL's speed is comparable to that of the typical dense retriever, DPR, as reported in the work~\cite{bevilacqua2022autoregressive}.
\begin{table*}[t] 
\centering 
\includegraphics[width=0.89\textwidth]{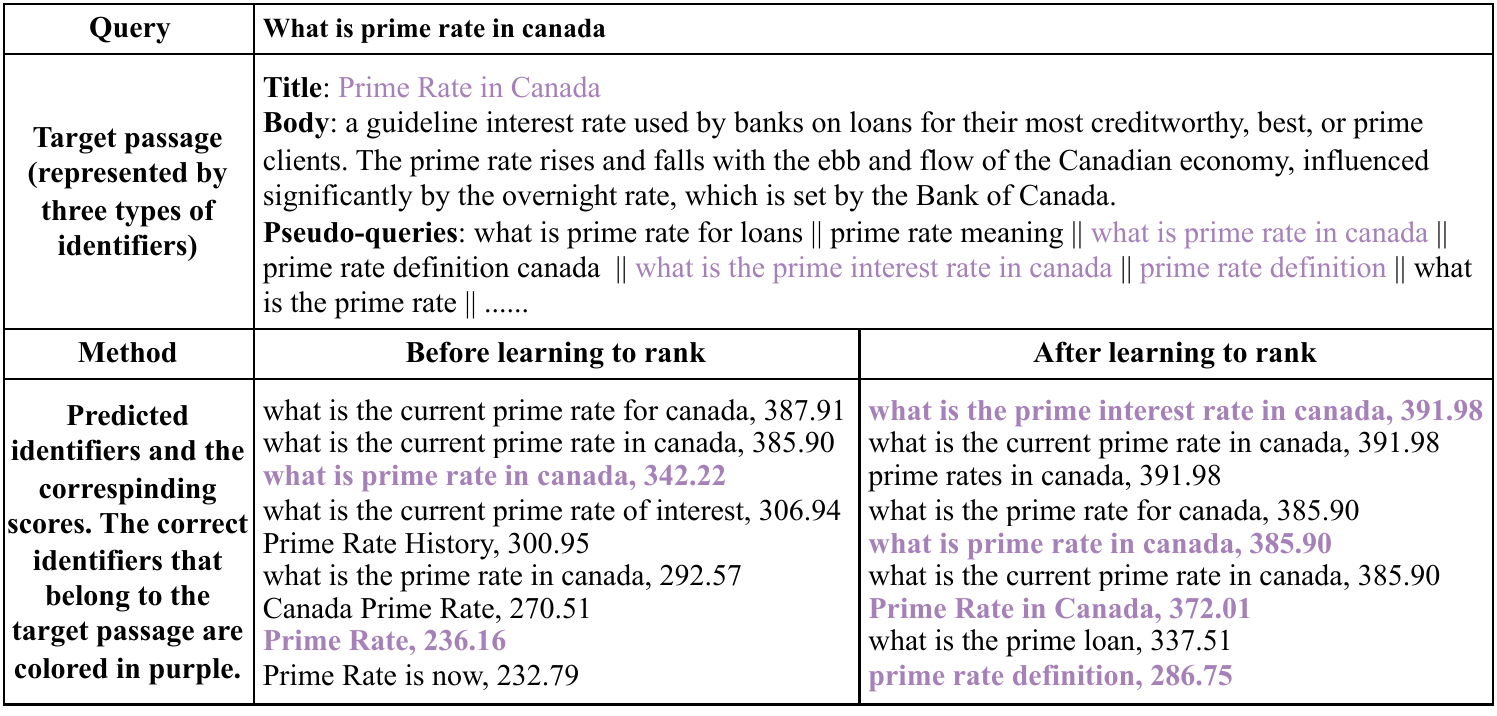} 
\vspace{-0.5em}
\caption{Case study on the MSMARCO dataset of the generative retrieval before and after learning to rank. The correctly predicted identifiers that belong to the target passage are colored in purple. }
\vspace{-1em}
\label{tab:case_study}
\end{table*}

\textbf{Margin analysis}. To assess the impact of margin values on retrieval performance, we manually set margin values ranging from 100 to 500 in Eq.~\ref{eqn4}. The results are summarized in Figure~\ref{margin_lambda_analysis:subfig1}. Our findings indicate that LTRGR with a margin of 100 performs worse than other variants, suggesting that a minimum margin value is necessary. As the margin value increases from 200 to 500, performance improves slightly but not significantly. While a larger margin can help the model better differentiate between positive and negative passages, it can also make the learning objective hard to reach.

\textbf{$\lambda$ analysis}. In the loss function described by Equation~\ref{eqn5}, we use a weight $\lambda$ to balance the contribution of the generation loss $\mathcal{L}_{gen}$ and the rank loss $\mathcal{L}_{rank}$. To determine the optimal weight values, we conducted a tuning experiment with different $\lambda$ values, and the results are summarized in Figure~\ref{margin_lambda_analysis:subfig2}. Our analysis yielded the following insights: 
1) Setting the weight to 0 leads to a significant performance gap, which confirms the importance of the generation loss, as discussed in Section 4.6. 
2) Varying the weight value from 500 to 200 has little effect on the performance in terms of hits@100, but the performance gradually decreases for hits@5 and hits@20 as the weight of the generation loss increases. This suggests that a higher weight of the generation loss can interfere with the function of the rank loss, which typically affects the top-ranking results such as hits@5 and hits@20.

\begin{figure}[t!]
\centering
\subfigure {
  \includegraphics[width=0.5\linewidth]{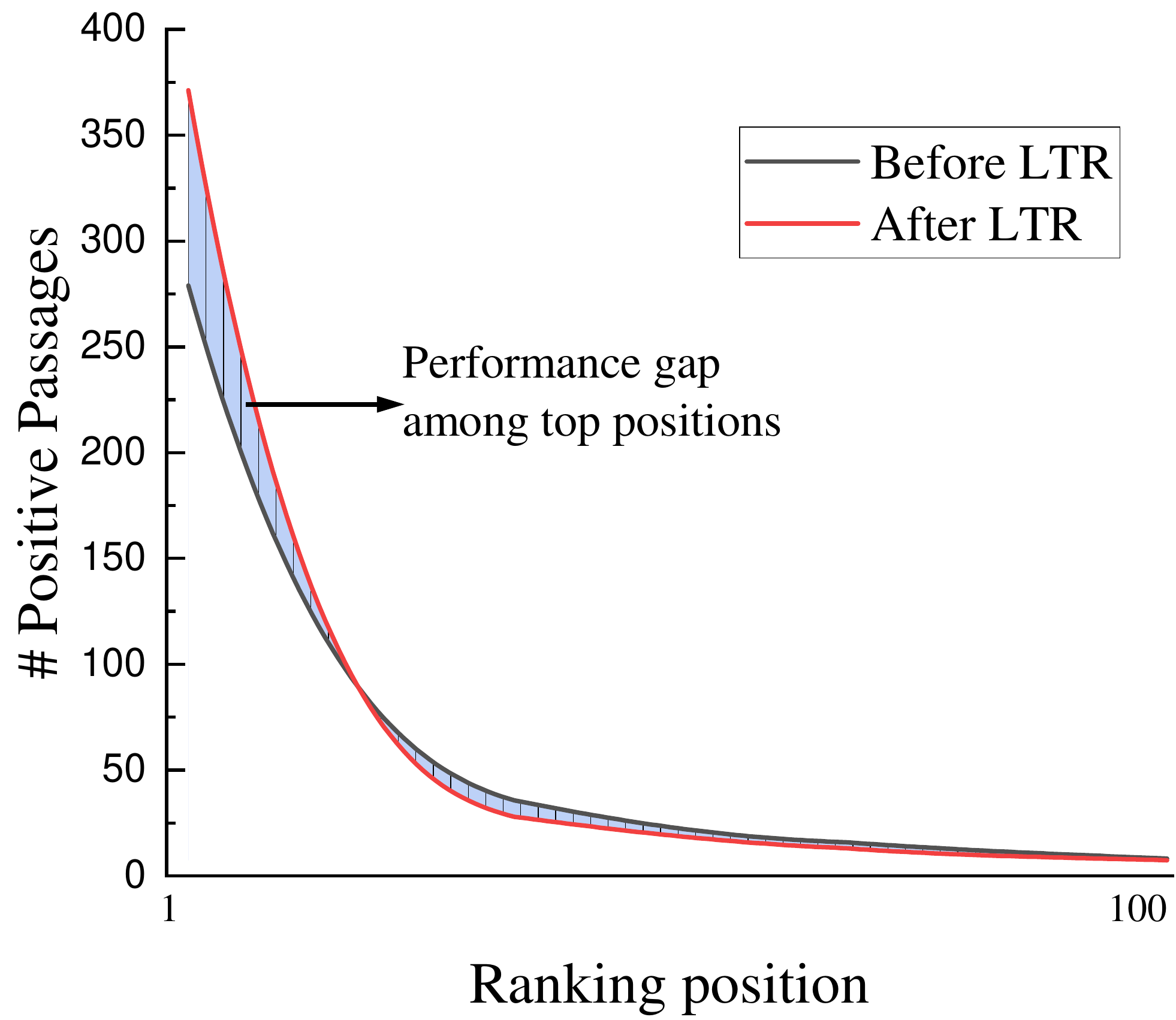}
   \label{Effective_LTR:subfig1}
   }
\vspace{-1em}
\caption{ The distribution of the number of retrieved positive passages is plotted against the ranking position on the MSMARCO dataset. The labels ``Before LTR'' and ``After LTR'' represent the generative model without and with learning-to-rank training, respectively.}
\vspace{-1em}
\label{Effective_LTR}
\end{figure}

\subsection{Effectiveness Analysis of Learning to Rank}
To better illustrate how the LTRGR works and what causes the performance improvement, we performed quantitative analysis and qualitative analysis (case study).

\textbf{Quantitative analysis}. We plotted the distribution of positive passages against their ranking positions in Figure~\ref{Effective_LTR:subfig1}. We used generative retrieval models before and after the learning-to-rank training to retrieve the top 100 passages from the MSMARCO dataset. We then counted the number of positive passages in each rank position in the retrieval list. By analyzing the results, we found that the performance improvement after the learning-to-rank training mainly comes from the top positions. LTRGR seems to push the positive passages to top-rank positions in the passage rank list. This vividly reflects the function of the rank loss $\mathcal{L}_{rank}$, which brings a better passage rank order to the list.

\textbf{Case Study}. To qualitatively illustrate the efficacy of the LTRGR framework, we analyzed the prediction results on MSMARCO in Table~\ref{tab:case_study}. It is observed that the number of the correct predicted identifiers gets  increased after the learning-to-rank training phase. Besides, for the same predicted identifier, such as ``what is prime rate in Canada'' in the case, its corresponding score also gets augmented after the learning-to-rank training. This clearly illustrates the effectiveness of the proposed learning-to-rank framework in generative retrieval, which enhances the autoregressive model to predict more correct identifiers with bigger corresponding scores. 

\section{Conclusion}
In this study, we introduce LTRGR, a novel framework that enhances current generative systems by enabling them to learn to rank passages. LTRGR requires only an additional training step via a passage rank loss and does not impose any additional burden on the inference stage. Importantly, LTRGR bridges the generative retrieval paradigm and the classical learning-to-rank paradigm, providing ample opportunities for further research in this field. Our experiments demonstrate that LTRGR outperforms other generative retrieval methods on three commonly used datasets. Moving forward, we anticipate that further research that deeply integrates these two paradigms will continue to advance generative retrieval in this direction.
\section*{Acknowledgments}
The work described in this paper was supported by Research Grants Council of Hong Kong (PolyU/5210919, PolyU/15207821, and PolyU/15207122), National Natural Science Foundation of China (62076212) and PolyU internal grants (ZVQ0).
\bibliography{aaai24}
\end{document}